\documentclass[runningheads]{llncs}
\usepackage[T1]{fontenc}
\usepackage{graphicx}
\usepackage{booktabs}
\usepackage[misc]{ifsym}
\usepackage{amsmath}
\usepackage{algorithm}
\usepackage{algorithmic}
\usepackage{multirow}
\usepackage{tabularx}
\usepackage{array}

\usepackage{mwe}

\begin{document}

\title{TwinBI: An Agentic Digital Twin for Efficient Augmented Interactions with Business Intelligence Dashboards}

\titlerunning{TwinBI: An Agentic Digital Twin for BI Dashboards}
\author{Jisoo Jang \and Wen-Syan Li}
\institute{Graduate School of Data Science, Seoul National University \\ \email{simonjisu@snu.ac.kr \& wensyanli@snu.ac.kr}}


\maketitle              

\begin{abstract}
Business intelligence (BI) increasingly combines dashboard interaction with LLM-based assistance, but these two modes often fall out of sync during multi-step analysis. As users switch between direct dashboard manipulation and natural-language queries, it becomes difficult to preserve a consistent analytical state across filters, hierarchies, metrics, and chart context. We present TwinBI, an agentic digital-twin framework that couples an LLM-based agent system with an executable BI dashboard state. TwinBI unifies conversational interaction, dashboard manipulation, semantic grounding, and provenance tracking through a shared analytical state reconstructed from a unified interaction log. It also exposes artifacts such as schema views, SQL, logs, and an \texttt{/insights} command for state-grounded analytical summaries. We evaluate TwinBI in two complementary ways. In a controlled A/B benchmark with the same backbone agent, TwinBI improves exact-match accuracy from 43.3\% to 63.3\%, partial-credit accuracy from 48.3\% to 70.8\%, and substantially reduces timeout rate from 40.0\% to 10.0\% relative to Dashboard alone. In a usability study, participants benefited from the integrated dashboard-and-chat workflow, with high task accuracy, moderate workload, and favorable ratings for state-aware interaction mechanisms. These results suggest that TwinBI improves both agent-level analytical reliability and user-facing analytical support by turning visible dashboard state into richer actionable context. Our dataset and source code are available at:

\url{https://github.com/simonjisu/TwinBI}

\end{abstract}

\section{Introduction}\label{sec:intro}

Business Intelligence (BI) systems form the core infrastructure that underpins data-driven decision-making in modern organizations. They allow analysts and decision-makers to investigate structured data, track organizational performance, and ground their decisions in measurable evidence~\cite{bi_system}. Recent progress in natural language processing, particularly in LLM-based agent architectures, has introduced a new interaction paradigm for BI. These systems are often presented as potential successors to traditional dashboards and analytics tools, converting natural language requests into tool executions and structured query language (SQL) statements.

Yet this emerging replacement narrative overlooks a long-standing disconnect between fluent natural language generation and analytically sound decision support. Enterprise business intelligence (BI) is grounded in precisely defined semantics—such as metric definitions, time assumptions, aggregation grains, and filter scopes—that are often encoded only implicitly in dashboards and semantic layers. LLM-based agents can drift outside these constraints, yielding answers that read well but are analytically inconsistent with the system's actual state. We therefore suggest that robust ``agentic BI'' may benefit from combining interactive BI tools with LLM-based assistance through an explicit coordination layer that aligns user intent, semantic definitions, and query execution, rather than relying on natural language alone as the interface.

To tackle this challenge, we present \textbf{TwinBI}, a framework that achieves BI by means of \textit{two interconnected digital twins}: an LLM agent twin that models user intent and reasoning and a BI twin that represents an executable analytics state, with both twins remaining synchronized throughout the interaction. TwinBI fuses natural-language interaction with machine-readable explicit representations of analytic schemas and hierarchies, metric dimension mappings, executable query specifications, and their associated result sets, while grounding the agent's behavior in the user's current analytical context, inferred from dashboard interactions.  
The system exposes intermediate analytical states (including tool invocations and query parameters) and captures complete provenance through unified event logging and persistent identifiers. This design promotes transparency and traceability for both user interactions and system-level reasoning. With this, BI-Twins shifts the role of LLM-based agents from ``replacing BI'' to ``working in concert with BI,'' thereby enhancing the robustness and reliability of decision support for business users.

In this paper, we present the design of TwinBI and evaluate it in two complementary ways. In a controlled A/B benchmark with the same backbone agent, TwinBI improves exact-match accuracy, partial-credit accuracy, and completion reliability over Dashboard alone. We further report a usability study showing that users benefit from the integrated dashboard-and-chat workflow for completing analysis tasks and interpreting results.


\section{Background}\label{sec:back}

Business Intelligence (BI) encompasses the methods, tools, and technologies that transform organizational data into actionable insights \cite{bi_system}. Many BI platforms rely on Online Analytical Processing (OLAP), where \textit{data cubes} organize multidimensional aggregates over \textit{measures} (e.g., sales or units sold) and \textit{dimensions} (e.g., time, geography, or product) \cite{gray1997data,chaudhuri1997overview}. Dimensions often contain hierarchies, such as Year $\succ$ Quarter $\succ$ Month, that support analysis at multiple granularities.

Analytical exploration over cubes is commonly described through operators such as \textit{slice}, \textit{dice}, \textit{roll up}, \textit{drill down}, and \textit{pivot}. In BI dashboards, these operators correspond to familiar actions such as filtering, changing time granularity, cross-filtering, changing group-by fields, and reconfiguring chart views, thereby forming the interaction vocabulary for navigating the underlying multidimensional space.

Large language models (LLMs) enable natural language interaction with data by converting user questions into structured representations and generating explanations grounded in retrieved evidence. Extending this idea, \textit{LLM agents} move beyond single-turn prompts to create tool-augmented workflows in which the agent decomposes a request, invokes external tools, and combines their outputs into a final response~\cite{fan2024survey,yao2022react}. 

A widely used strategy for building LLM-powered BI assistants is the \textit{Natural Language to SQL} (NL2SQL) pipeline, which converts a user's request into an SQL query, executes it, and returns the result in natural language~\cite{zhang2024natural,liu2024survey}. This approach is practical because it maps user intent directly to executable analytical queries over the underlying database.

Despite these advances, LLM agents still fall short of being full BI platforms, as operational analytics needs far more than SQL generation. In particular, BI systems must preserve analytical state across both natural-language interaction and direct dashboard manipulation while keeping metrics, filters, and hierarchy semantics aligned throughout multi-step exploration.

\begin{figure}[t]
  \centering
  \includegraphics[width=0.8\textwidth]{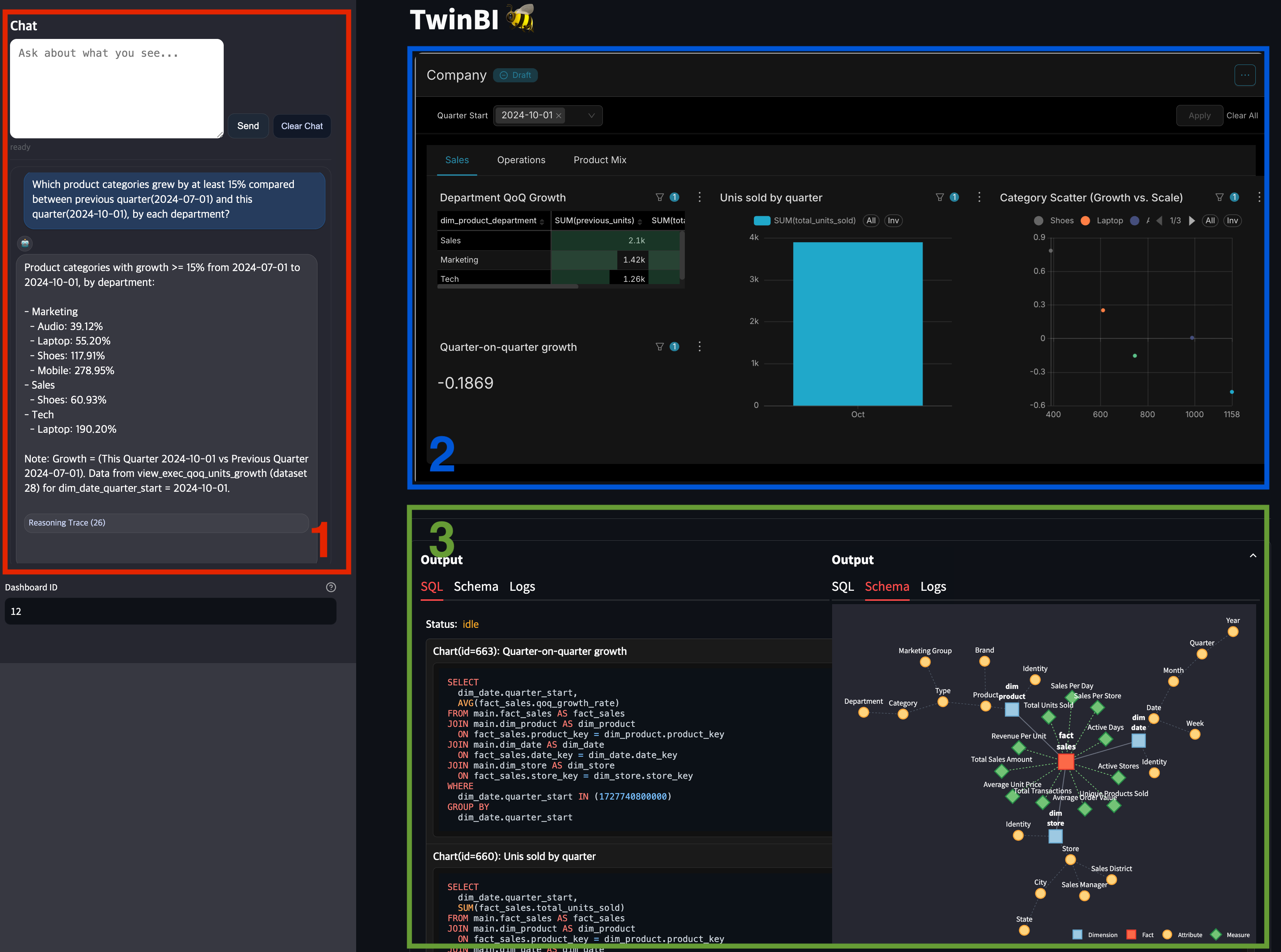}
   \caption{The TwinBI user interface: (1) a chat interface for natural-language analytics queries, (2) an embedded dashboard for interactive exploration, and (3) an inspection panel exposing artifacts such as SQL and the hierarchical schema to support schema understanding.}
  \label{fig:ui_all}
\end{figure}

\begin{figure}[t]
  \centering
  \includegraphics[width=0.75\textwidth]{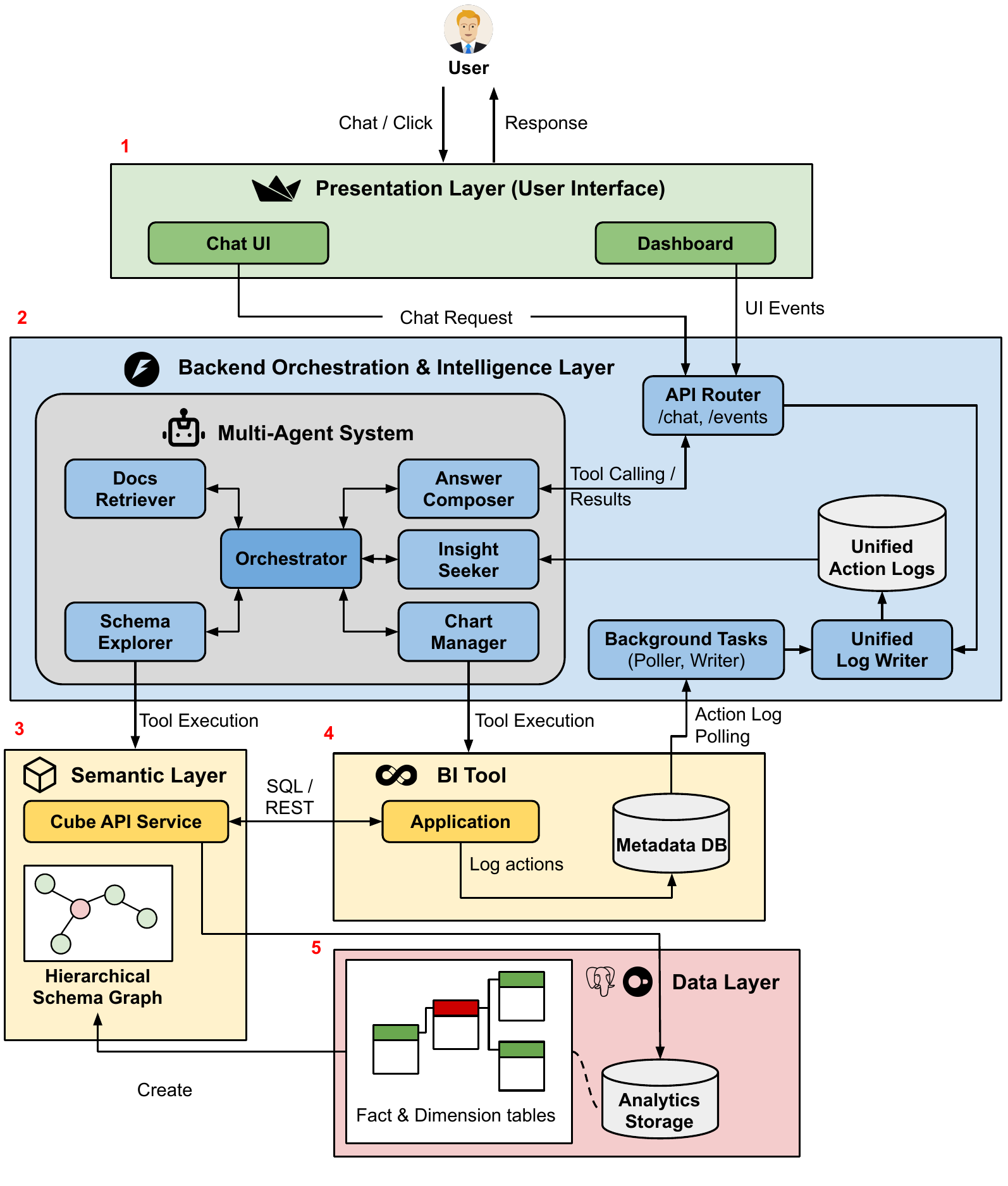}
   \caption{System Architecture of TwinBI.}
  \label{fig:arch}
\end{figure}

\section{System Architecture}\label{sec:system}

TwinBI adopts a layered architecture that synchronizes an LLM-based agent system with an executable BI dashboard state. The architecture preserves a consistent analytical state across conversational interaction and direct dashboard manipulation while maintaining end-to-end traceability through unified logs.

Figure \ref{fig:ui_all} shows the TwinBI interface, which combines natural-language interaction, embedded dashboards, and inspection views for schema- and query-level artifacts. Figure~\ref{fig:arch} illustrates the overall design of the system. The system is containerized using Docker \cite{docker_docs} to ensure isolated and reproducible deployment. The architecture consists of five layers: (1) Presentation Layer, (2) Orchestration Layer, (3) Semantic Layer, (4) BI Tool Layer, and (5) Data Layer.

\subsection{Presentation Layer}

The \textit{presentation layer} provides the user-facing experience, combining a chat interface with embedded BI dashboards. Users can submit natural-language queries or interact directly with visualizations via filtering, tab switching, cross-filtering, and drill-down operations.

To keep both modalities aligned, the interface tracks the active analytical context, including the selected chart, tab, and recent dashboard interactions, and sends these signals to the backend for state reconstruction.

The interface is built with Streamlit \cite{streamlit_docs} and incorporates Apache Superset \cite{superset_software} dashboards to provide interactive visualizations.

\begin{table}[t]
  \caption{Major user interface activities used for unified interaction logs.}
  \centering
  \renewcommand{\arraystretch}{1.2}
  \resizebox{0.9\textwidth}{!}{%
  \begin{tabular}{p{0.3\columnwidth} p{0.68\columnwidth}}
    \toprule
    \textbf{Category} & \textbf{Description} \\
    \midrule
    Tab navigation & Switches tabs and records the active tab. \\
    Series toggle & Toggles chart series visibility through legend interaction. \\
    Enable cross-filter & Applies a chart-driven cross-filter to linked views. \\
    Disable cross-filter & Removes a previously propagated cross-filter. \\
    Global filter insertion & Adds a dashboard-level filter. \\
    Global filter removal & Removes a dashboard-level filter. \\
    \bottomrule
  \end{tabular}%
  }

  \label{tab:ui_activity_taxonomy}
\end{table}

\subsection{Backend Orchestration \& Intelligence Layer}

The \textit{backend orchestration and intelligence layer} manages both the multi-LLM agent system and the executable dashboard state. Implemented with FastAPI \cite{fastapi_docs}, it combines recent dialogue history, dashboard interaction logs, and tool outputs into a unified analytical context, routes sub-tasks to specialized agents, and consolidates their results into responses anchored in the current state.

All interactions with external systems occur through backend-governed tools. The backend also maintains a \textit{unified interaction log} that captures conversational exchanges, dashboard operations, and tool metadata as the authoritative record for state reconstruction and provenance. Table~\ref{tab:ui_activity_taxonomy} summarizes the major dashboard interaction events recorded in this log.

\subsection{Semantic Layer}

The \textit{semantic layer} captures business meaning using declarative models of measures, dimensions, hierarchies, and join paths. It provides the shared semantic model for both conversational outputs and dashboard queries, enforcing compatible grains and valid joins. We also derive a Hierarchy Schema Graph from fact tables and dimension hierarchies, giving the Schema Explorer agent a structured and navigable view of the analytical schema.

This layer is built on top of Cube \cite{cube_core}, which provides REST and SQL interfaces for executing model-driven queries.

\subsection{BI Tool}
The \textit{BI tool layer} offers interactive dashboards for visual data exploration. TwinBI uses Apache Superset \cite{superset_software} both to render visualizations and to capture detailed interaction events, which are ingested into the unified log and replayed through the chart data API when contextual grounding is required.

\subsection{Data Layer}

The Data Layer serves as analytical storage for domain-specific data. Analytical datasets are stored in DuckDB \cite{duckdb_docs} and accessed solely through the semantic layer, preserving uniform metric definitions and aggregation behavior while keeping storage decoupled from interaction handling.

\section{Functionality of TwinBI System} 
\label{sec:function}

TwinBI is designed for the common BI workflow in which users alternate between clicking through an existing dashboard and asking follow-up questions in natural language. Instead of treating these as separate modes, the system reuses the dashboard state accumulated through interaction and applies it to subsequent chat requests. In practice, this means that charts, filters, and follow-up questions are resolved against the same reconstructed analytical state rather than against an isolated prompt.

\begin{figure}[t]
  \centering
  \includegraphics[width=\textwidth]{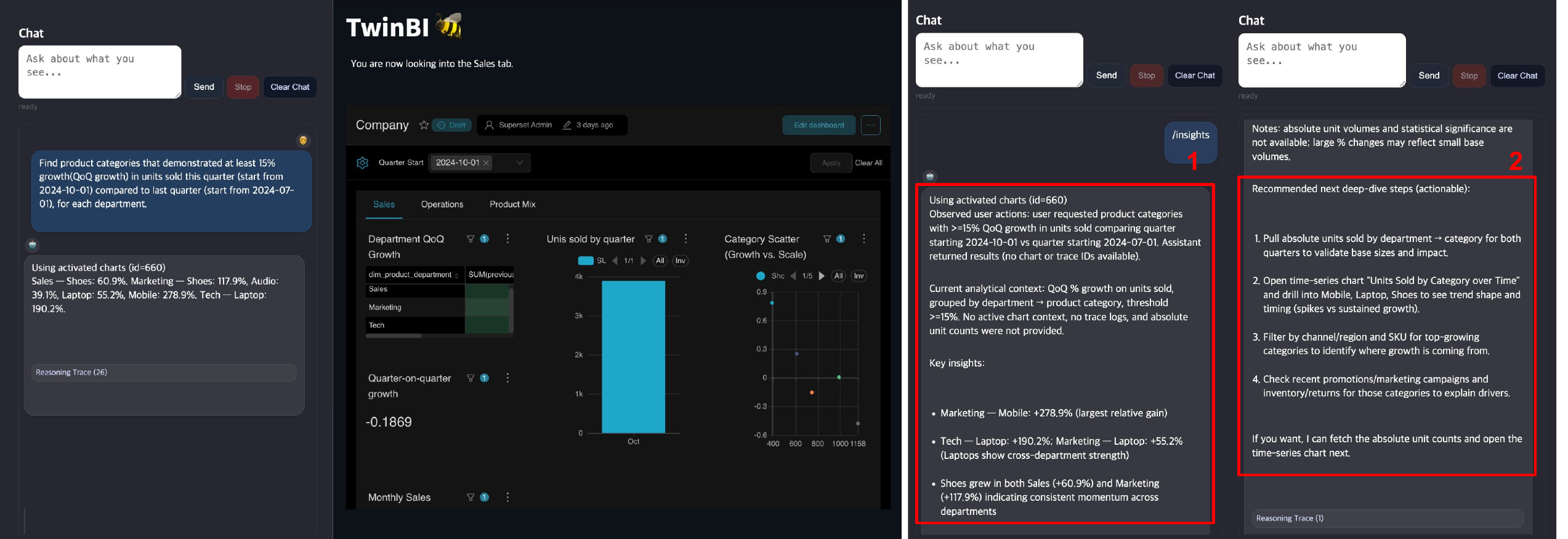}
  \caption{(Left) Example query requesting product categories with more than 15\% quarter-over-quarter unit sales growth in the fourth quarter. (Right) Example \texttt{/insights} output summarizing high-growth product categories, caveats about percentage-based interpretation, and suggested next analysis steps.}
  \label{fig:functionality}
\end{figure}

\subsection{Finding and Creating Charts}

Users can access or generate charts through two complementary mechanisms: (1) direct interaction with the BI dashboard and (2) natural-language prompts via the chat interface. 

In dashboard-centric interaction, users navigate and refine existing visualizations by applying filters, switching tabs, using cross-filtering, drilling down, and toggling series visibility. We explicitly log these actions because the resulting state is often not recoverable from a later chat turn alone. For example, a follow-up question such as ``Why did this category increase?'' is underspecified unless the system can recover which chart was active, which filters were already applied, and which hierarchy level the user had reached. TwinBI therefore encodes dashboard actions as structured events and uses them to reconstruct the current analytical state before interpreting a new conversational request.

In conversational interaction, users can request new visualizations without needing to specify schema names or rebuild the dashboard context from scratch. The system resolves the request through the semantic layer and keeps any generated chart aligned with the current slice of analysis. We found this important in cases where users wanted to branch from an existing dashboard view rather than start a new query from a blank state.

\subsection{Hierarchy Schema Graph, SQL, and Logs}

TwinBI exposes three inspection artifacts for users who want to verify what the system is doing rather than accept a final answer at face value. First, the Hierarchy Schema Graph gives a compact view of measures, dimensions, and hierarchies through the Schema Explorer. This is mainly useful when a user knows the business concept they want but not the exact field names supported by the semantic model. Second, TwinBI exposes the SQL associated with each chart query so that users can inspect joins, filters, and aggregation choices. Third, the unified interaction log can be inspected directly to trace how a conversational answer relates to earlier dashboard operations and tool calls. We include these artifacts because debugging BI answers often requires checking whether an error came from schema selection, filter carry-over, or answer generation.

\subsection{Finding Insights by Agent}

As shown on the right side of Figure \ref{fig:functionality},
TwinBI offers a dedicated \texttt{/insights} command for moments when users want a state-aware summary of the current view rather than an answer to a new question. When this command is executed, the backend assembles a compact execution context from the unified interaction log, including recent conversations, tool traces, the active chart, and its current filters. This context is passed to a specialized insight agent that returns a short summary organized around three elements: the current analytical slice, the main quantitative observations visible in that slice, and sensible next checks for the user.

This function is intentionally constrained so summary-style outputs remain grounded in the currently visible analytical evidence. It only summarizes information supported by the current analytical state and must indicate when evidence is insufficient for a stronger claim. Figure~\ref{fig:functionality} illustrates this by showing a summary tied to the current dashboard context rather than to new exploratory queries.

\section{Experiments}\label{sec:exper}

We design a benchmark-style A/B evaluation to measure whether TwinBI's state-grounded orchestration improves analytical task completion over a Dashboard system under matched model and environment conditions. Unlike the usability study in Section~\ref{sec:usability}, this experiment targets controlled agent performance on a fixed query set and focuses on exact-match accuracy, robustness, and interaction efficiency.

\subsection{Experimental Setting}

The evaluation uses a retail sales dashboard environment built over a shared semantic model with product, store, and date as its primary analytical dimensions. In both experimental conditions, we employ a Playwright-based browser agent configured with \texttt{gpt-5-mini} as the decision-making model \cite{playwright,openaiIntroducingGPT5}, subject to a maximum budget of 30 interaction steps. At each step, the agent observes the current state, selects the next action, executes it through Playwright, and then updates its next decision from the newly observed state. Here, the state includes the current screenshot, actionable UI candidates, recent action history, and task-specific prompt context. We compare two systems: \textbf{(A)} Dashboard, which makes these stepwise decisions from the visible dashboard alone, and \textbf{(B)} TwinBI, which makes the same kind of stepwise decisions but augments them with chat interface and backend support with \texttt{gpt-5-mini} for Orchestration \& Intelligence Layer.

The benchmark consists of 30 analytical queries. The query set is balanced across five task families, with six queries in each family: (1) store and district ranking, (2) premium product analysis, (3) quarter-over-quarter growth analysis, (4) comparison and aggregation tasks across dashboard views, and (5) robustness and trap tasks that test policy compliance and filter stability. 

To construct the target answers used for evaluation, we resolved each query through three independent paths: direct database queries, cube-API queries, and dashboard-level queries. We then checked these three answers for self-consistency and performed a final manual verification step before fixing the reference annotation for each benchmark item.


\subsection{Evaluation Metrics}

We evaluate our system using both outcome-oriented and behavior-oriented metrics. The primary outcome measures are: (1) \textit{Exact match accuracy}, which assesses whether the final structured prediction is identical to the reference annotation; (2) \textit{Partial-credit accuracy}, which quantifies slot-level correctness for partially accurate structured outputs, thereby enabling us to distinguish near-miss reasoning from complete failure; and (3) \textit{Average steps} to completion, which operationalizes interaction efficiency as the total number of recorded steps divided by the number of queries.

The behavioral metrics are defined as follows: (1) \textit{Timeout rate} captures the proportion of queries that terminate upon reaching the maximum step budget instead of producing a valid answer. (2) \textit{Invalid action rate} denotes the proportion of recorded interaction steps that either violate the prescribed action policy or reference an interface element that is unusable. (3) \textit{Loop query rate} represents the proportion of queries that exhibit consecutive repeated action signatures, whereas \textit{loop step rate} denotes the proportion of all recorded steps that are part of such repetitive loops. For example, the metric counts repeated chat steps only when the same chat prompt is issued in consecutive turns, and repeated click steps only when the same click target coordinates recur consecutively.

\begin{table}[t]
  \caption{Aggregate A/B results with \texttt{gpt-5-mini} fixed across both conditions over the 30-query benchmark set. Average steps are computed from the total recorded steps in each run.}
  \label{tab:ab_main_results}
  \centering
  \renewcommand{\arraystretch}{1.11}
  \setlength{\tabcolsep}{5pt}
  \resizebox{0.5\textwidth}{!}{%
  \begin{tabular}{lccc}
    \toprule
    \textbf{System} & \textbf{Exact} & \textbf{Partial} & \textbf{Avg. Steps} \\
    \midrule
    Dashboard & 43.33\% & 48.33\% & 16.47 \\
    TwinBI & 63.33\% & 70.83\% & 6.90 \\
    \bottomrule
  \end{tabular}
  }
\end{table}

\begin{figure}[t]
  \centering
  \includegraphics[width=\textwidth]{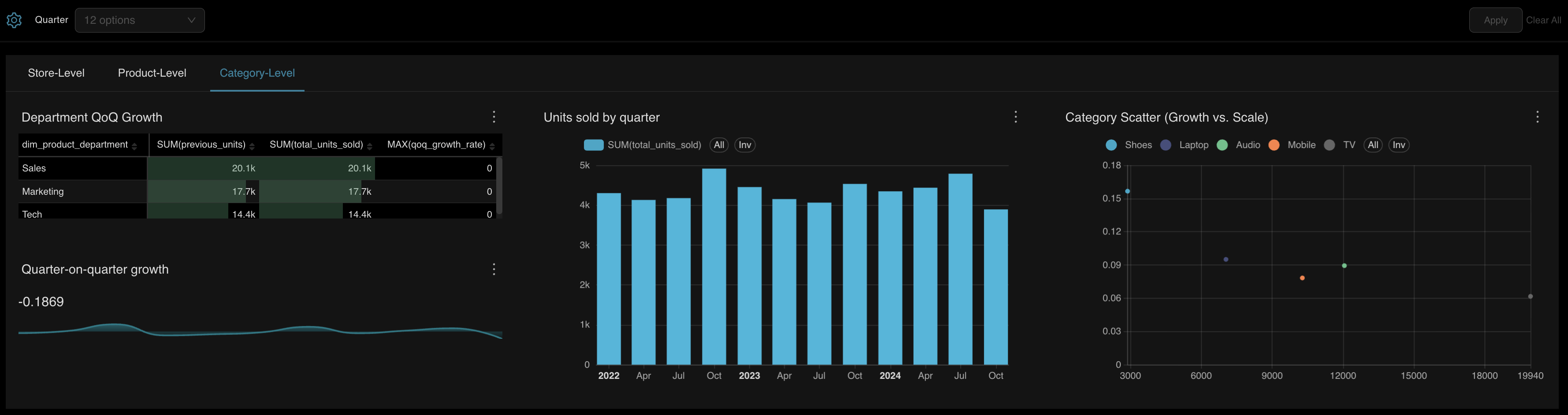}
  \caption{Category-level dashboard view with \textit{Department QoQ Growth} (left), \textit{Units sold by quarter} (center), and \textit{Category Scatter (Growth vs.\ Scale)} (right).}
  \label{fig:result1_dashboard}
\end{figure}

\subsection{Results}

Table~\ref{tab:ab_main_results} summarizes the aggregate A/B results. TwinBI improved both exact-match and partial-credit accuracy while requiring fewer steps per query. Figure~\ref{fig:result1_dashboard} shows the category-level dashboard view used in the representative Q14 example. This gain appears to come from enriching visible dashboard state with structured context through the backend agent and chat interface, rather than relying on dashboard probing alone. Q14 is a \textit{QoQ slot-filling} task that asks for the department--category pair with the highest quarter-over-quarter unit sales growth from Q3 to Q4. To solve it from Dashboard alone, the system must apply the \texttt{2024-10-01} filter, select \texttt{Marketing} in the \textit{Department QoQ Growth} table to trigger cross-filtering, and then hover over \texttt{Mobile} in the \textit{Category Scatter (Growth vs.\ Scale)} chart to recover the required tuple. Dashboard localized this view but then spent the full 30-step budget on repeated hover-based probing without completing this chain reliably, whereas TwinBI opened the same view and recovered the answer through a single chat query grounded in the visible dashboard state. 

\begin{table}[t]
  \caption{Behavior-oriented metrics from the same A/B run. TwinBI reduced timeouts and invalid actions substantially, although its loop-step rate remained non-trivial because some failures still involved repeated chat-centered reasoning patterns.}
  \label{tab:ab_failure_results}
  \centering
  \renewcommand{\arraystretch}{1.15}
  \setlength{\tabcolsep}{4pt}
  \resizebox{0.8\textwidth}{!}{%
  \begin{tabular}{lcccc}
    \toprule
    \textbf{System} & \textbf{Timeout} & \textbf{Invalid Action} & \textbf{Loop Query} & \textbf{Loop Step} \\
    \midrule
    Dashboard & 40.00\% & 10.93\% & 36.67\% & 29.76\% \\
    TwinBI & 10.00\% & 0.00\% & 27.59\% & 39.13\% \\
    \bottomrule
  \end{tabular}
  }
\end{table}

Table~\ref{tab:ab_failure_results} reports the corresponding behavior metrics. The lower timeout and invalid-action rates indicate that TwinBI makes the interaction process more stable, not only more accurate. This robustness gain appears to be associated with replacing brittle UI-only probing with chat-supported interpretation over richer structured context; Q17 is a representative case. The task asks for the Q3-to-Q4 QoQ growth rate of the \texttt{Mobile} category in the \texttt{Marketing} department. In the Dashboard trace, the system reached the correct category-level view and attempted the right interaction chain, but then remained stuck in repeated hover-based tooltip probing and late filter switching to recover the final value. In contrast, TwinBI opened the same view and issued a chat query over the visible dashboard, which returned the required QoQ growth rate directly and allowed the run to terminate successfully. The loop metrics suggest a more mixed shift in failure mode: TwinBI reduced repeated-failure queries, but some remaining failures still involved repeated chat-centered steps rather than dashboard-level probing.

This A/B test is intended to isolate the effect of system support for the same backbone agent. The results suggest that TwinBI improves completion reliability and structured interpretation by turning visible dashboard state into richer actionable context under the same agent and step budget.



\section{Usability Study}\label{sec:usability}

We evaluate how TwinBI supports users as they move between dashboard-based exploration and conversational follow-up. Rather than measuring open-ended long-term adoption, this study focuses on whether users can complete representative analytical tasks, how much interaction effort they expend, which interaction patterns they prefer, and whether the system helps them articulate correct higher-level interpretations.

\subsection{Methodology}

We conducted a within-subjects usability study with five participants. Each participant completed three analytical scenarios with progressively increasing analytical complexity and system support.

In each scenario, participants had to locate entities that met predefined performance constraints using multidimensional filtering and aggregation. The scenarios varied in how many interaction modalities were available, enabling us to disentangle the effects of orchestration and analytical state grounding.

\begin{itemize}
    \item\textbf{S1: Store Performance Analysis (Limited Support).}
    Participants identified the top-performing store in the North district based on average daily sales, excluding stores with fewer than 15 active days per month. Dashboard filtering and chart inspection were available; chat assistance was optional.
    
    \item\textbf{S2: Product Pricing Analysis (Moderate Support).}
    Participants identified product types whose average revenue per unit exceeded the overall portfolio average. The interface provided dashboard interaction and conversational chart requests after initial dashboard interaction.
    
    \item\textbf{S3: Category Growth Analysis (Full Support).}
    Participants identified categories that achieved a minimum of 15\% growth in unit sales quarter over quarter between Q3 (beginning on 2024-07-01) and Q4 (beginning on 2024-10-01). All interaction mechanisms, including conversational chart generation and insight support, were available.
     
\end{itemize}

This staged design lets us observe not only whether participants succeed, but also how their behavior changes as more state-aware assistance becomes available. In particular, we were interested in whether participants would continue relying on direct dashboard interaction, shift toward chat once context was established, or combine the two modes.

For evaluation, we report both objective and subjective measures. The objective measures are: (1) \textit{Task Accuracy}, indicating whether participants solved each scenario correctly; (2) \textit{Interaction Cost}, measured as the number of dashboard events and chat turns per scenario; and (3) \textit{Insight Accuracy}, indicating whether participants produced correct higher-level interpretations. The subjective measures are: (1) \textit{Perceived Difficulty} on a 5-point Likert scale; (2) \textit{Feature Usefulness}, covering dashboard interaction, chart finding, click+chat, chat-only interaction, SQL inspection, schema exploration, interaction log inspection, and \texttt{/insights}; and (3) \textit{NASA-TLX} dimensions for mental demand, temporal demand, performance, effort, and frustration \cite{hart1988nasa_tlx}.

\begin{table}[t]
  \caption{Summary of scenario-level results (S1--S3). Accuracy metrics are reported as percentages. Interaction cost represents the mean number of dashboard clicks and chat turns per scenario. Perceived difficulty reflects participants' average self-reported task difficulty on a 5-point scale.}
  \centering
  \renewcommand{\arraystretch}{1.1}
  \small
  \setlength{\tabcolsep}{6pt}
  \begin{tabular}{lccc}
    \toprule
    \textbf{Metric} & \textbf{S1} & \textbf{S2} & \textbf{S3} \\
    \midrule
    Task Accuracy (\%) & 100\% & 73.33\% & 100\% \\
    Insight Accuracy (\%) & 80\% & 100\% & 80\% \\
    Average Interaction Cost (Clicks) & 6.4 & 34 & 49 \\
    Average Interaction Cost (Chats) & 0.6 & 6 & 5.2 \\
    Average Perceived Difficulty (1--5) & 1.8 & 3.4 & 4.2 \\
    \bottomrule
  \end{tabular}
  \label{tab:quant_results}
\end{table}

\begin{figure}[t]
  \centering
  \includegraphics[width=0.8\textwidth]{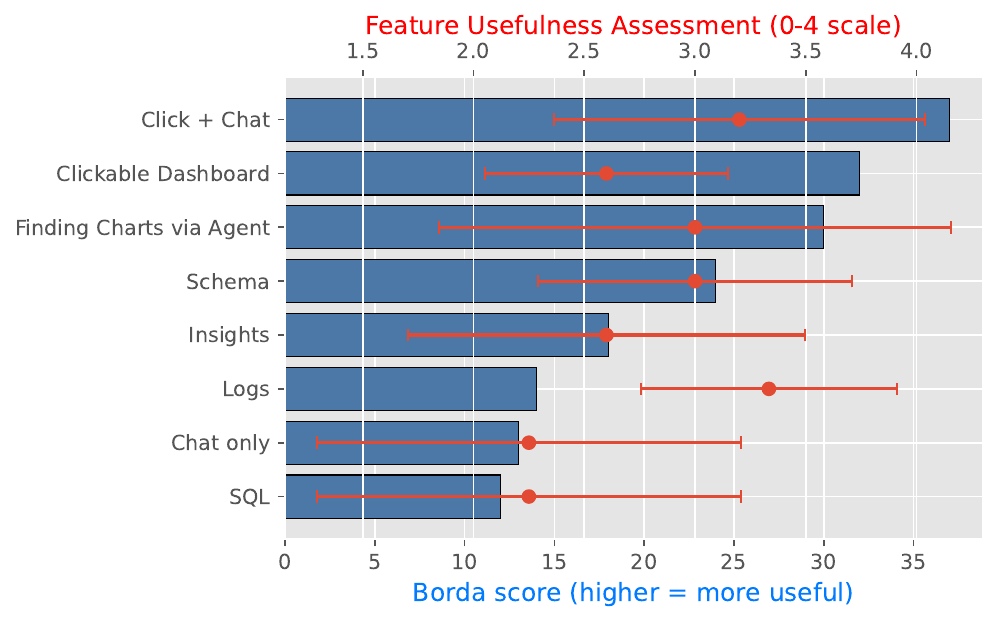}
  \caption{Feature usefulness combining relative preference (Borda scores) and usefulness ratings (mean with standard deviation). Bars show Borda scores; points show feature usefulness assessment ratings (0 responses treated as N/A).}
  \label{fig:feature_useful}
\end{figure}

\begin{figure}[t]
\centering 
\includegraphics[width=0.8\textwidth]{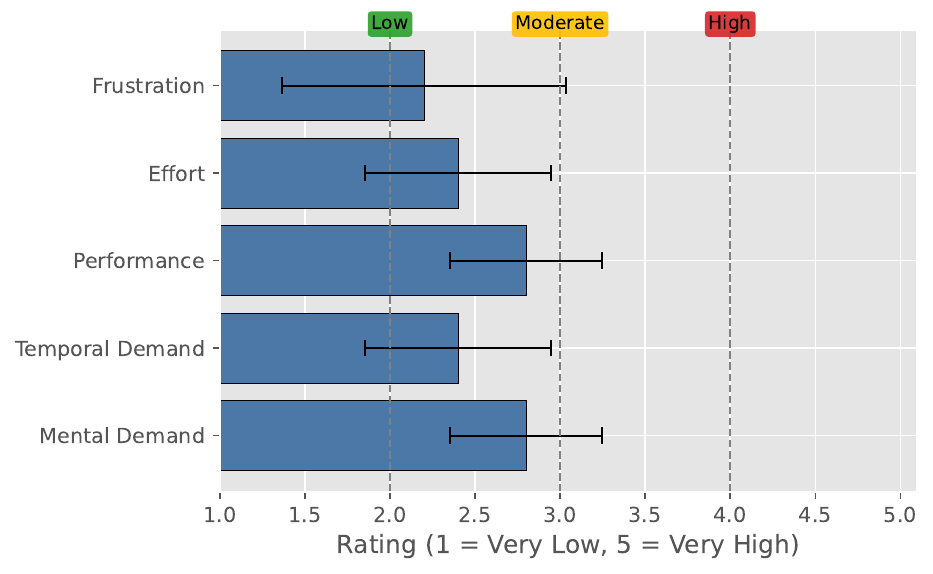}
\caption{NASA-TLX workload ratings ($N=5$). Bars represent mean scores with standard deviation error bars. Dashed vertical lines indicate workload thresholds (Low, Moderate, High), color-coded for interpretability.}
\label{fig:NASA}
\end{figure}

\subsection{Results}

\noindent\textbf{RQ1: To what extent does TwinBI simplify BI workflows and minimize operational friction without compromising analytical accuracy?}

Table~\ref{tab:quant_results} reports the scenario-level outcomes. Task Accuracy remained high across all scenarios, although the harder scenarios required substantially more interaction. Participants completed S1 with little assistance and relatively few clicks, whereas S2 and S3 led to heavier use of both dashboard interaction and chat. In the study, participants usually used the dashboard to establish context first and then turned to chat for comparison, threshold checking, or explanation across multiple views. Figure~\ref{fig:NASA} shows that workload still stayed in the low-to-moderate range despite this increase in interaction cost. Given the small sample, we interpret this result as evidence that the combined workflow is usable for moderately complex tasks rather than as proof of broad efficiency gains.

\vspace{5pt}
\noindent\textbf{RQ2: Does analytical state awareness improve agent assistance effectiveness?}

Figure~\ref{fig:feature_useful} presents the perceived usefulness results. Participants consistently ranked state-aware combinations such as the Clickable Dashboard, Finding Charts via Agent, and Click+Chat above chat-only interaction and direct SQL inspection. During the study, participants rarely abandoned the dashboard entirely. More often, they located the relevant view first and then used chat to clarify or summarize what they were already seeing. Borda aggregation, together with inter-participant agreement (Kendall's $W=0.62$, $p<0.01$), shows that this ordering was fairly stable across participants \cite{emerson2013original,kendall1939problem}.

Chat use increased in the more complex scenarios, especially when participants needed help interpreting a filtered view rather than finding it. In that sense, the benefit came less from replacing dashboard interaction and more from supporting follow-up interpretation once the dashboard context was already in place.

\vspace{5pt}
\noindent\textbf{RQ3: Does the unified interaction log improve insight generation and reflective reasoning?}

Insight Accuracy was 80\% in S1, 100\% in S2, and 80\% in S3, so most participants were able to produce correct higher-level interpretations in the guided scenarios. Direct use of \texttt{/insights}, however, was limited: three of the five participants invoked it, and two of those three produced fully correct insights. That usage pattern is too small to support a strong claim, but it does suggest that the feature can be useful when participants have already narrowed the analysis to a specific view.

The usefulness ratings in Figure~\ref{fig:feature_useful} point in the same direction. Participants responded more positively to log-derived support when it appeared through higher-level tools than when the raw interaction log was shown directly. In this study, provenance seemed most helpful when it reduced interpretation work rather than when it became another object of inspection.

\begin{table}[t]
\caption{Representative functionality comparison of closely related systems. Type: C = commercial, P/S = paper/system. Y = explicitly supported; P = partially supported or unclear; -- = not stated or not supported.}
\centering
\footnotesize
\setlength{\tabcolsep}{4pt}
\renewcommand{\arraystretch}{1.1}
\resizebox{\textwidth}{!}{%
\begin{tabular}{p{3.8cm} c c c c c c c}
\toprule
\textbf{System} & \textbf{Type} &
\textbf{NLQ} &
\textbf{Dashboard} &
\textbf{Sync} &
\textbf{Dash Ops} &
\textbf{Schema} &
\textbf{Logging} \\
\midrule
Power BI Copilot~\cite{powerbi_copilot_intro} & C & Y & Y & P & Y & Y & P \\
Amazon QuickSight~\cite{amazonq_quicksight_ga_2024} & C & Y & Y & -- & Y & -- & -- \\
Tableau Einstein~\cite{tableau_einstein_copilot_release} & C & Y & Y & P & Y & Y & P \\
Looker~\cite{looker_conversational_analytics_ga} & C & Y & Y & P & Y & Y & P \\
LIDA~\cite{dibia2023lida} & P/S & Y & -- & -- & P & -- & -- \\
Chat2VIS~\cite{maddigan2023chat2vis} & P/S & Y & -- & -- & P & -- & -- \\
WaitGPT~\cite{xie2024waitgpt} & P/S & Y & -- & -- & -- & -- & -- \\
InsightLens~\cite{weng2025insightlens} & P/S & Y & -- & -- & -- & -- & -- \\
Hey Dashboard!~\cite{dhanoa2025hey} & P/S & Y & Y & P & Y & -- & -- \\
TwinBI (ours) & P/S & Y & Y & Y & Y & Y & Y \\
\bottomrule

\end{tabular}
}%
\label{tab:system-comparison-functionalities}
\end{table}

\section{Related Works}
\label{sec:related}
Natural language interfaces to data (NLIDB), NL-to-SQL systems, and recent LLM agents lower the barrier to querying structured data \cite{zhang2024natural,liu2024survey,yao2022react,openai_agents}. In BI settings, however, the central issue is not only query generation but also preserving semantic consistency across metrics, filter scope, aggregation grain, and iterative interaction with dashboards \cite{bi_system}. Table~\ref{tab:system-comparison-functionalities} therefore compares representative prior work along the capabilities most relevant to this setting: natural-language querying, dashboard support, synchronization between chat and dashboard state, dashboard operations, schema support, and logging.

Prior academic systems mainly address partial slices of this space, such as NL-driven chart generation, chat-oriented analytical assistance, or dashboard usability, rather than synchronized state management across conversational and dashboard interaction \cite{dibia2023lida,maddigan2023chat2vis,xie2024waitgpt,weng2025insightlens,dhanoa2025hey}.

Commercial BI copilots increasingly combine NLQ with dashboards and semantic layers \cite{powerbi_copilot_intro,amazonq_quicksight_ga_2024,tableau_einstein_copilot_release,looker_conversational_analytics_ga}. However, public documentation still suggests only partial support for explicit synchronization, schema-grounded interaction continuity, or comprehensive provenance logging. TwinBI fills this gap by combining dashboard interaction, conversational querying, explicit synchronization, schema-aware reasoning, and unified logging in one system.

\section{Conclusion}\label{sec:conclusion}
We presented \textbf{TwinBI}, an agentic digital-twin framework that unifies conversational interaction and direct dashboard manipulation through a shared analytical state. TwinBI combines semantic grounding, dashboard state reconstruction, and unified provenance tracking so that users and agents can operate over the same analytical context. Across a controlled A/B benchmark and a usability study, our results suggest that this design improves analytical reliability, interaction stability, and user-facing analytical support over dashboard interaction alone.

Future work includes testing on larger datasets and more diverse users, improving chart grounding and value extraction for complex analyses, extending TwinBI to transfer analytical state across dashboards, and exploring how TwinBI can better support agentic decision-making workflows.

\vfill
\bibliographystyle{splncs04}
\bibliography{00_references}

@misc{bi_system,
  author       = {{Tableau Software, LLC}},
  title        = {{Business intelligence: A complete overview}},
  year         = {2025},
  howpublished = {https://www.tableau.com/business-intelligence/what-is-business-intelligence},
  note         = {Accessed: 2026-01-22}
}

@misc{streamlit_docs,
  author       = {{Streamlit, Inc.}},
  title        = {Streamlit Documentation},
  howpublished = {https://docs.streamlit.io/},
  year         = {2018},
  note         = {Accessed: 2026-01-22}
}

@misc{playwright,
  author       = {{Microsoft}},
  title        = {Playwright},
  howpublished = {https://github.com/microsoft/playwright},
  year         = {2020},
  note         = {Accessed: 2026-01-22}
}

@misc{openaiIntroducingGPT5,
	author = {OpenAI},
	title = {{I}ntroducing {G}{P}{T}-5},
	howpublished = {https://openai.com/index/introducing-gpt-5/},
	year = {2025},
	note = {Accessed: 2026-01-22},
}

@software{superset_software,
  author       = {{Apache Software Foundation}},
  title        = {Apache Superset},
  howpublished = {GitHub repository},
  url          = {https://github.com/apache/superset},
  year         = {2016},
  note         = {Accessed: 2026-01-22}
}

@misc{fastapi_docs,
  author       = {{FastAPI}},
  title        = {FastAPI Documentation},
  howpublished = {https://fastapi.tiangolo.com/},
  year         = {2018},
  note         = {Accessed: 2026-01-22}
}

@misc{docker_docs,
  author       = {{Docker, Inc.}},
  title        = {Docker Documentation},
  howpublished = {https://docs.docker.com/},
  year         = {2013},
  note         = {Accessed: 2026-01-22}
}

@misc{duckdb_docs,
  author       = {{DuckDB Labs}},
  title        = {DuckDB Documentation},
  howpublished = {https://duckdb.org/docs/stable/},
  year         = {2024},
  note         = {Accessed: 2026-01-22}
}

@misc{openai_agents,
  author       = {{OpenAI}},
  title        = {New tools for building agents},
  howpublished = {https://openai.com/index/new-tools-for-building-agents/},
  year         = {2025},
  note         = {Accessed: 2026-01-22}
}

@misc{powerbi_copilot_intro,
  author       = {{Microsoft}},
  title        = {Copilot in Power BI: Introduction},
  howpublished = {Microsoft Learn},
  url          = {https://learn.microsoft.com/en-us/power-bi/create-reports/copilot-introduction},
  note         = {Accessed: 2026-02-11}
}

@misc{amazonq_quicksight_ga_2024,
  author       = {{Amazon}},
  title        = {Amazon Q is Now Generally Available in Amazon QuickSight},
  howpublished = {AWS Business Intelligence Blog},
  year         = {2024},
  url          = {https://aws.amazon.com/blogs/business-intelligence/amazon-q-is-now-generally-available-in-amazon-quicksight-bringing-generative-bi-capabilities-to-the-entire-organization/},
  note         = {Accessed: 2026-02-11}
}

@misc{tableau_einstein_copilot_release,
  author       = {{Tableau}},
  title        = {Tableau Release: Einstein Copilot and Related Features},
  howpublished = {Tableau Blog},
  url          = {https://www.tableau.com/blog/release-tableau-einstein-copilot-tableau-multi-fact-relationships-viz-extensions},
  note         = {Accessed: 2026-02-11}
}

@misc{looker_conversational_analytics_ga,
  author       = {{Gemini}},
  title        = {Looker Conversational Analytics is Now Generally Available},
  howpublished = {Google Cloud Blog},
  url          = {https://cloud.google.com/blog/products/business-intelligence/looker-conversational-analytics-now-ga},
  note         = {Accessed: 2026-02-11}
}

@software{cube_core,
  author       = {{Cube Contributors}},
  title        = {Cube Core},
  howpublished = {GitHub repository},
  url          = {https://github.com/cube-js/cube},
  year         = {2019},
  note         = {Accessed: 2026-01-22}
}

@inproceedings{yao2022react,
  title={React: Synergizing reasoning and acting in language models},
  author={Yao, Shunyu and Zhao, Jeffrey and Yu, Dian and Du, Nan and Shafran, Izhak and Narasimhan, Karthik R and Cao, Yuan},
  booktitle={The eleventh international conference on learning representations},
  year={2022}
}

@article{liu2024survey,
  title={A Survey of NL2SQL with Large Language Models: Where are we, and where are we going?},
  author={Liu, Xinyu and Shen, Shuyu and Li, Boyan and Ma, Peixian and Jiang, Runzhi and Zhang, Yuxin and Fan, Ju and Li, Guoliang and Tang, Nan and Luo, Yuyu},
  journal={arXiv preprint arXiv:2408.05109},
  year={2024}
}

@article{gray1997data,
  title={Data cube: A relational aggregation operator generalizing group-by, cross-tab, and sub-totals},
  author={Gray, Jim and Chaudhuri, Surajit and Bosworth, Adam and Layman, Andrew and Reichart, Don and Venkatrao, Murali and Pellow, Frank and Pirahesh, Hamid},
  journal={Data mining and knowledge discovery},
  volume={1},
  number={1},
  pages={29--53},
  year={1997},
  publisher={Springer}
}

@article{chaudhuri1997overview,
  title={An overview of data warehousing and OLAP technology},
  author={Chaudhuri, Surajit and Dayal, Umeshwar},
  journal={ACM Sigmod record},
  volume={26},
  number={1},
  pages={65--74},
  year={1997},
  publisher={ACM New York, NY, USA}
}

@article{zhang2024natural,
  title={Natural language interfaces for tabular data querying and visualization: A survey},
  author={Zhang, Weixu and Wang, Yifei and Song, Yuanfeng and Wei, Victor Junqiu and Tian, Yuxing and Qi, Yiyan and Chan, Jonathan H and Wong, Raymond Chi-Wing and Yang, Haiqin},
  journal={IEEE Transactions on Knowledge and Data Engineering},
  volume={36},
  number={11},
  pages={6699--6718},
  year={2024},
  publisher={IEEE}
}

@inproceedings{fan2024survey,
  title={A survey on rag meeting llms: Towards retrieval-augmented large language models},
  author={Fan, Wenqi and Ding, Yujuan and Ning, Liangbo and Wang, Shijie and Li, Hengyun and Yin, Dawei and Chua, Tat-Seng and Li, Qing},
  booktitle={Proceedings of the 30th ACM SIGKDD conference on knowledge discovery and data mining},
  pages={6491--6501},
  year={2024}
}

@article{dhanoa2025hey,
  title={Hey Dashboard!: Supporting Voice, Text, and Pointing Modalities in Dashboard Onboarding},
  author={Dhanoa, Vaishali and Le{\'o}n, Gabriela Molina and Hoggan, Eve and Gr{\"o}ller, Eduard and Streit, Marc and Elmqvist, Niklas},
  journal={arXiv preprint arXiv:2510.12386},
  year={2025}
}

@article{weng2025insightlens,
  title={InsightLens: Augmenting LLM-Powered Data Analysis with Interactive Insight Management and Navigation},
  author={Weng, Luoxuan and Wang, Xingbo and Lu, Junyu and Feng, Yingchaojie and Liu, Yihan and Feng, Haozhe and Huang, Danqing and Chen, Wei},
  journal={IEEE Transactions on Visualization and Computer Graphics},
  year={2025},
  publisher={IEEE}
}

@inproceedings{xie2024waitgpt,
  title={Waitgpt: Monitoring and steering conversational llm agent in data analysis with on-the-fly code visualization},
  author={Xie, Liwenhan and Zheng, Chengbo and Xia, Haijun and Qu, Huamin and Zhu-Tian, Chen},
  booktitle={Proceedings of the 37th Annual ACM Symposium on User Interface Software and Technology},
  pages={1--14},
  year={2024}
}

@article{dibia2023lida,
  title={LIDA: A tool for automatic generation of grammar-agnostic visualizations and infographics using large language models},
  author={Dibia, Victor},
  journal={arXiv preprint arXiv:2303.02927},
  year={2023}
}

@article{maddigan2023chat2vis,
  title={Chat2vis: Generating data visualizations via natural language using chatgpt, codex and gpt-3 large language models},
  author={Maddigan, Paula and Susnjak, Teo},
  journal={Ieee Access},
  volume={11},
  pages={45181--45193},
  year={2023},
  publisher={Ieee}
}

@incollection{hart1988nasa_tlx,
  author    = {Hart, Sandra G. and Staveland, Lowell E.},
  title     = {Development of NASA-TLX (Task Load Index): Results of Empirical and Theoretical Research},
  booktitle = {Human Mental Workload},
  editor    = {Hancock, Peter A. and Meshkati, Najmedin},
  series    = {Advances in Psychology},
  volume    = {52},
  pages     = {139--183},
  year      = {1988},
  publisher = {North-Holland}
}

@article{kendall1939problem,
  title={The problem of m rankings},
  author={Kendall, Maurice G and Smith, B Babington},
  journal={The annals of mathematical statistics},
  volume={10},
  number={3},
  pages={275--287},
  year={1939},
  publisher={JSTOR}
}

@article{emerson2013original,
  title={The original Borda count and partial voting},
  author={Emerson, Peter},
  journal={Social Choice and Welfare},
  volume={40},
  number={2},
  pages={353--358},
  year={2013},
  publisher={Springer}
}


\end{document}